\documentclass{article}

\usepackage[sort, numbers]{natbib}
\setcitestyle{square}

 \usepackage[final]{nips_2016}

\usepackage[utf8]{inputenc} 
\usepackage[T1]{fontenc}    
\usepackage{hyperref}       
\usepackage{url}            
\usepackage{booktabs}       
\usepackage{amsfonts}       
\usepackage{nicefrac}       
\usepackage{microtype}      

\title{Ranking Biomarkers Through Mutual Information}

\usepackage{amsmath}
\usepackage{amsthm}
\usepackage{mathtools}
\usepackage{cancel}
\usepackage{mathtools}
\def\Ypos{y=1}
\def\X{\boldsymbol{X}}
\def\x{\boldsymbol{x}}

\newcommand{\argmax}[1]{\underset{#1}{\operatorname{arg}\,\operatorname{max}}\;}

\theoremstyle{definition}

\theoremstyle{theorem}
\newtheorem{theorem}{Theorem}

\usepackage{caption}
\usepackage{subcaption}

\usepackage{wrapfig}

\author{
Konstantinos Sechidis\\
School of Computer Science\\
University of Manchester\\
\texttt{konstantinos.sechidis@manchester.ac.uk} \\
\And
Emily Turner \\
School of Computer Science\\
University of Manchester\\
\texttt{emily.turner@manchester.ac.uk} \\
\And
Paul D. Metcalfe\\
Advanced Analytics Centre\\
Global Medicines Development, AstraZeneca\\
\texttt{paul.metcalfe@astrazeneca.com}\\
\And
James Weatherall\\
Advanced Analytics Centre,\\
Global Medicines Development, AstraZeneca\\
\texttt{james.weatherall@astrazeneca.com}\\
\And
Gavin Brown \\
School of Computer Science\\
University of Manchester\\
\texttt{gavin.brown@manchester.ac.uk} \\
}

\begin{document}

\maketitle

\begin{abstract}
We study information theoretic methods for ranking biomarkers. In clinical trials there are two, closely  related, types of biomarkers: predictive and prognostic, and disentangling them is a key challenge. Our first step is to phrase biomarker ranking in terms of optimizing an information theoretic quantity. This formalization of the problem will enable us to derive rankings of predictive/prognostic biomarkers, by estimating different, high dimensional, {\em conditional mutual information} terms. To estimate these terms, we suggest efficient low dimensional approximations, and we derive an empirical Bayes estimator, which is suitable for small or sparse datasets. Finally, we introduce a new visualisation tool that captures the {\em prognostic} and the {\em predictive} strength of a set of biomarkers. We believe this representation will prove to be a powerful tool in biomarker discovery.
\end{abstract}

\section{Introduction}
We present an information theoretic approach to disentangle predictive and prognostic biomarkers. 
In clinical trials, a {\em prognostic biomarker} is a clinical or biological characteristic that provides information on the likely outcome irrespective of the treatment. On the other hand a {\em predictive biomarker}, is a clinical or biological characteristic that provides information on the likely benefit from treatment. One of the key challenges in personalised medicine is to discover predictive biomarkers which will guide the analysis for tailored therapies, while discovering prognostic biomarkers is crucial for general patient care \citep{Ruberg2015}. We should clarify that our work focuses on hypothesis generation (exploratory analysis), instead of hypothesis testing (confirmatory analysis) \citep{DmitrienkoEtAll2016}.

In our work we will focus on a clinical dataset $\mathcal{D}=\{y_i,\x_i,t_i\}_{i=1}^n$, where, $y$ is a realization of a binary target variable $Y,$ $t$ is a realization of binary treatment indicator $T$ (i.e. $T=1$ if patient received experimental treatment, $0$ otherwise), and $\x$ is a $p$-dimensional realization of the feature vector $\X,$ which describes the joint random variable of the $p$ categorical features (or biomarkers). To make the distinction between prognostic and predictive biomarkers more formal we will follow a strategy introduced by various previous works \citep{FosterEtAll2011,LipkovichDmitrienko2014b}. Let us assume that the true underlying model is the following logistic regression with up to second order interaction terms:
{
\begin{align}
\text{logit}P(\Ypos|t,\x) = \alpha + \sum_{i=1}^p \beta_{i} x_{i} + \sum_{i,j=1}^p \beta_{i,j} x_{i}x_{j}
+ \gamma t  + \left( \sum_{i=1}^p \delta_{i} x_{i} + \sum_{i,j=1}^p \delta_{i,j} x_{i}x_{j} \right)t. \notag
\end{align}
}
Covariates with non-zero $\beta$ coefficients are prognostic, while the ones with non-zero $\delta$ coefficients are  predictive.
Our work proposes an information theoretic framework for deriving two different rankings of the biomarkers, one that captures their {\em prognostic} strength, and one that captures their {\em predictive} strength. On top of that, we introduce a visualisation tool  that captures both the {\em prognosticness} and the {\em predictiveness} of a set of biomarkers. This tool enables us to identify potentially undiscovered biomarkers, worthy of further investigation.

\section{Background on Biomarker Ranking}
\label{sec:Back_BiomarkerRanking}
This section connects the problem of biomarker discovery, in context of the machine learning problem of feature selection and the clinical trials problem of subgroup identification.
\subsection{Prognostic Biomarker Discovery and Feature Selection}
\label{sec:Back_Prognostic}
We now demonstrate that the problem of selecting {\em prognostic biomarkers} is equivalent to feature selection using a supervised dataset  $\{y_i,\x_i\}_{i=1}^n$. There are many different methods for feature selection, but we will focus on information theoretic approaches, where, firstly we {\em rank} the features and then we {\em select} the top-$k$ ones that contain most of the useful information. 
The underlying objective function is to find the smallest feature set $\X^*$ that maximizes $I(\X^*;Y)$, or in other words that the shared information between $\X^*$ and $Y$ is maximized:
\begin{align}
\X^* = \argmax{\X_{\theta} \in \X} {I}(\X_{\theta};Y). \notag
\end{align}
\citet{BrownPocockZhaoLujan2012}  derived a greedy optimization process which assesses features based on a simple scoring criterion on the utility of including a feature. At each step we select the feature $X_k$ that maximizes the conditional mutual information (CMI): $J^{\text{CMI}}(X_k) = \hat{I}(X_k;Y|{\bf{X}}_{\theta}),$ where ${\bf{X}}_{\theta}$ is the set of the features already selected. As the number of selected features grows, the dimension of $X_{\theta}$ also grows, and this makes our estimates less reliable. To overcome this problem {\em low order} criteria have been derived. For example, by ranking the features independently on their mutual information with the class, we derive a ranking that takes into account the \emph{relevancy} with the class label. Choosing the features according to this ranking corresponds to the {\em Mutual Information Maximization} (MIM) criterion; where the score of each feature $X_k$ is given by: $J^{\text{MIM}}(X_k)=I(X_k;Y).$ This approach does not take into account the \emph{redundancy} between the features. By using more advanced techniques  \citep{PengLongDing2005}, we can take into account both the relevancy  and the redundancy between the features themselves, {\em without} having to compute very high dimensional distributions. \citet{BrownPocockZhaoLujan2012}  showed that a criterion that controls relevancy, redundancy, conditional redundancy and provides a very good tradeoff in terms of accuracy, stability and flexibility is the {\em Joint Mutual Information} (JMI) criterion \citep{YangMoody1999}: $J^{\text{JMI}}(X_k) = \sum_{X_j \in {\bf{X}}_{\theta}}\hat{I}(X_k;Y|X_j).$ Through heuristic, this guarantees to increase the likelihood at each step.

While the above framework has been suggested for supervised scenarios, our aim is to explore how it can been extended to be useful in clinical trial scenarios, i.e. $\mathcal{D}$. The extra treatment variable $T$ provides interesting dynamics, but before showing our suggested extension,  we will briefly present the literature on predictive biomarkers and subgroup identification.

\subsection{Predictive Biomarker Discovery and Subgroup Identification}
\label{sec:Back_Predictive}
The problem of deriving {\em predictive biomarkers} is closely related to the problem of subgroup identification \citep{DmitrienkoEtAll2016}. In clinical trials, patient populations cannot be considered homogeneous, and thus the effect of treatment will vary across different subgroups of the population. Exploring the heterogeneity of subject responses to treatment is very critical for drug development, which is underlined by a draft Food and Drug Administration guidance \citep{Ruberg2015}. As a result consideration of patient subgroups is necessary in multiple stages of trial development. 
\citet{Berry1990} gives the following definition: subgrouping is a partition of the set of all patients into disjoint subsets or subgroups and it is usually determined by a small number of measurable covariates, which are the predictive biomarkers. In the traditional subgroup identification problem the set of predictive biomarkers is relatively small, i.e. 2-3 biomarkers \citep{LipkovichEtAll2011}.

In the literature there are many different methods for subgroup identification. A popular one is {\em recursive partitioning} of the covariate space, using criteria that capture the interaction between $T$ and $Y$ \citep{SuEtAll2009,LipkovichEtAll2011,LohEtAll2015}. Another solution builds upon the {\em counterfactual modelling} idea: firstly by deriving a new variable for each patient that captures the treatment effect and then using this variable to select or rank the covariates. For example, \citet{FosterEtAll2011} can be seen as exploring the covariate space which maximizes the odds-ratio between $T$ and $Y$. In the following section, we will show that starting from a natural objective function, we can  derive predictive biomarkers by exploring areas that maximize the mutual information between $T$ and $Y$.

\section{An Information Theoretic View on Biomarker Ranking}
Our work extends the feature ranking framework from supervised to clinical trial data. The treatment variable $T$ provides extra useful information, and a natural way to capture this is by the following criterion: to maximize the shared mutual information between the target $Y$ and the joint random variable of the treatment $T$ and the optimal feature set $\X^*$, or in information theoretic notation:
$
\X^* = \textrm{argmax}~I (\X_{\theta}T;Y).
$
By using the chain rule \citep{CoverThomas2006}, these objective can be decomposed as follows in the following way:
{
\begin{align}
\X^* = \argmax{\X_{\theta} \in \X} {I}(\X_{\theta}T;Y) =\argmax{\X_{\theta} \in \X} \Big( \underbrace{{I}(\X_{\theta};Y)}_{\mathclap{\text{Prognostic term}}} + \underbrace{{I}(T;Y|\X_{\theta})}_{\mathclap{\text{Predictive term}}}\Big) \notag
\end{align}
}
The first term, captures the features with prognostic power, while the second captures the features with predictive power. By optimizing these two terms independently we can derive two different objectives for the two different features set:
$\X^*_{\text{Prog}} = \argmax{\X_{\theta} \in \X} {I}(\X_{\theta};Y)$ and $\X^*_{\text{Pred}} = \argmax{\X_{\theta} \in \X} {I}(T;Y|\X_{\theta}).$ Similar to \citep{BrownPocockZhaoLujan2012}, to optimize these two objectives, we can derive a greedy optimization process, where are each step we select the feature $X_k$ that maximizes the following terms:
\begin{align}
J_{{Prog}}(X_k)=I(X_k;Y|\X_{{Prog}}),~~~~~~~~~~~~~~~~J_{{Pred}}(X_k)=I(T;Y|X_k\X_{{Pred}}). \notag
\end{align}

where ${\bf{X}}_{{Prog}}$ are the features already been ranked as prognostic, while  ${\bf{X}}_{{Pred}}$ as predictive. As the number of selected features grows, the dimension of ${\bf{X}}_{{Prog}}$ and ${\bf{X}}_{{Pred}}$ also grows, and this makes the estimates less reliable. To overcome this issue, we derive low-order approximations, such as the one presented in Section \ref{sec:Back_Prognostic}.
\subsection{Lower-order approximations}
With the following theorem we present our main contribution -- lower order approximations of $J_{\text{Prog}}(X_k)$ and $J_{\text{Pred}}(X_k)$:

\begin{theorem}
\label{thm}
The first two order approximations for deriving Prog. and Pred. rankings are given by: \\
\begin{minipage}{0.46\linewidth}
{
\begin{align}
 J^{1^{st}}_{\text{Prog}}(X_k) & = I(X_k;Y), \notag \\
J^{2^{nd}}_{\text{Prog}}(X_k) & = \sum_{X_j \in \X_{\text{Prog}}}{I}(X_k;Y|X_j), \notag
\end{align}
}
\end{minipage}
\ \ \
\begin{minipage}{0.52\linewidth}
{
\begin{align}
J^{1^{st}}_{\text{Pred}}(X_k) & = I(T;Y|X_k). \notag \\
J^{2^{nd}}_{\text{Pred}}(X_k) & = \sum_{X_j \in \X_{\text{Pred}}}{I}(T;Y|X_kX_j). \notag
\end{align}
}
\end{minipage}\par\vspace{\belowdisplayskip}

{\rm \noindent Proof sketches: For prognostic, the proof is identical to \citep{BrownPocockZhaoLujan2012}, while for the predictive we can prove these approximations by combining the results of \citet{BrownPocockZhaoLujan2012} with the chain rule \citep{CoverThomas2006}.}
\end{theorem}

For example, by making assumptions similar to the ones of MIM, we can derive the $1^{st}$-order criteria for deriving prognostic and predictive rankings respectively. These criteria do not take into account interactions between features, and as a result fail to capture the {\em redundancy}. To overcome this limitation so we can use higher order criteria, such as JMI, which explores $2^{nd}$-order interaction terms between features.\newpage
\subsection{Estimating Conditional Mutual Information Through an Empirical Bayes Approach}
\begin{wrapfigure}{r}{0.35\textwidth}
\vskip-0.5cm
           \includegraphics[width=0.35\textwidth]{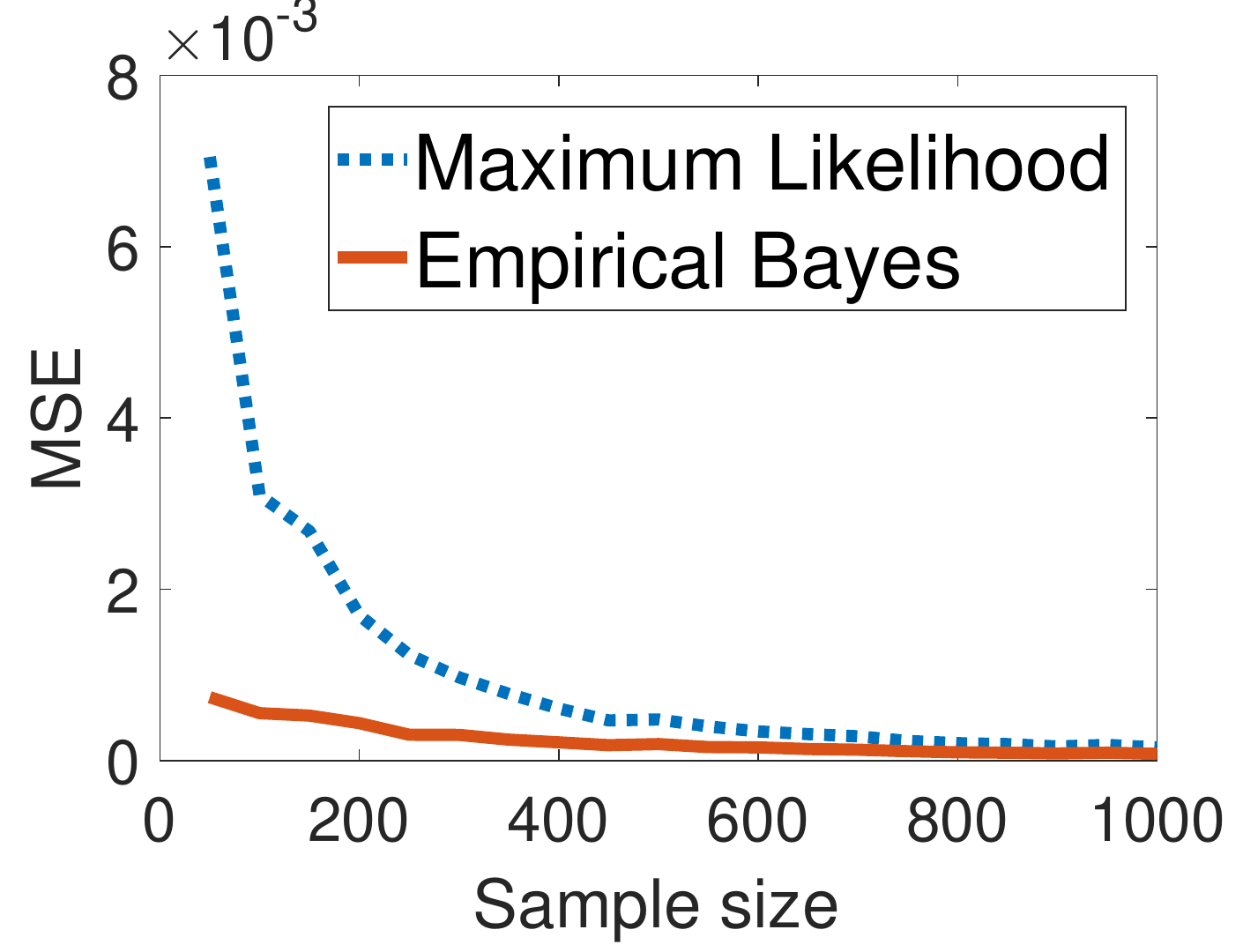}
       \caption{MSE between maximum likelihood and empirical Bayes estimator for conditional mutual information $I(T;Y|X)=0$ with $|\mathcal{T}|=|\mathcal{Y}|=2$ and $|\mathcal{X}|=25$.}
           \label{fig:MSE}

\end{wrapfigure}
In order to derive the above rankings we need to estimate conditional mutual information terms. In our work we will focus on categorical data, and we derive an efficient way for estimating these terms through an empirical Bayes procedure. Due to space limitations we omit the technical details, but our analysis extends a work on entropy estimation. \citet{hausser2009} suggested an entropy estimator that employs James-Stein-type shrinkage at the level of cell frequencies.  Building upon this, we derived an estimator for the conditional mutual information. Our proposed estimator achieves smaller mean squared error than maximum-likelihood, especially in ``small $n$, large $p$'' scenarios -- which are common in micro-array data. For example, Figure \ref{fig:MSE} compares the performance of the maximum likelihood estimator against our proposed empirical bayes approach, and as we observe our proposed estimator converges much faster.



\section{Predictive--Prognostic (PP) Graphs}
We now present a visualisation tool  that captures both the {\em prognostic} and the {\em predictive} power of a set of biomarkers (PP-graphs). We believe that this representation will provide useful information over both the prognostic and predictive power of each biomarker, and it will be helpful for controlling false discoveries in clinical trials. For example, in subgroup-identification (Section \ref{sec:Back_Predictive}), we define interesting subgroupings by using predictive biomarkers. Many methods, such as the counterfactual modelling, i.e. Virtual-twins suggested by \citep{FosterEtAll2011}, derive as predictive, biomarkers that are strongly prognostic. Using a PP-graph we get more insight over the prognostic and predictive power of each biomarker and this may help in eliminating these type of errors.

Now we will show these graphs through a motivating example.
We will use the same data generation model as in \citep{FosterEtAll2011}.  Let us assume that we simulate randomized trials with $1000$ patients, and the $X$s are generated as independent $X_j \sim N(0,1),j=1...15$. We consider logit models for data generation
\[ \text{logit} P(\Ypos|t,\x) = -1 + 0.5x_1 + 0.5x_2 -0.5x_7  + 0.5x_2x_7 + 0.1 t + 1.5 t \mathbb{I}(x_1>0 \cap x_2<0 \cap x_3>0).\]
Thus, the patients with $\left( x_1>0 \cap x_2<0 \cap x_3>0 \right)$ will have an enhanced treatment effect. As a result the three variables, $X_1, X_2$ and $X_3$, are the predictive biomarkers. Furthermore, $X_1, X_2$ and $X_7$ are the three prognostic biomarkers and the other nine biomarkers are irrelevant. 

\begin{figure}[h]

    \centering
    \begin{subfigure}[b]{0.32\textwidth}
        \includegraphics[width=\textwidth]{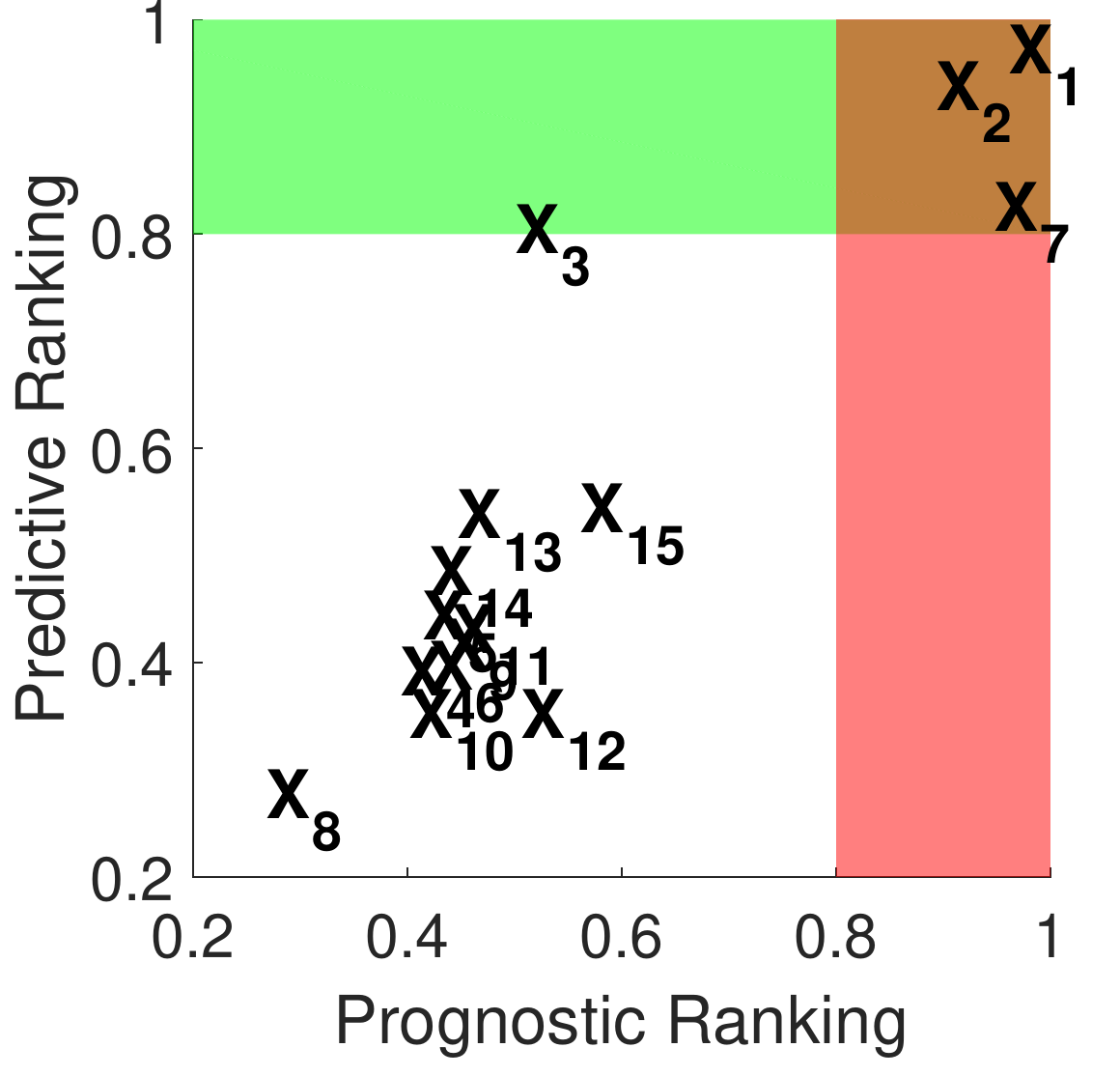}
        \caption{Virtual-twins.}
        \label{fig:Counter}
    \end{subfigure}
    \begin{subfigure}[b]{0.32\textwidth}
        \includegraphics[width=\textwidth]{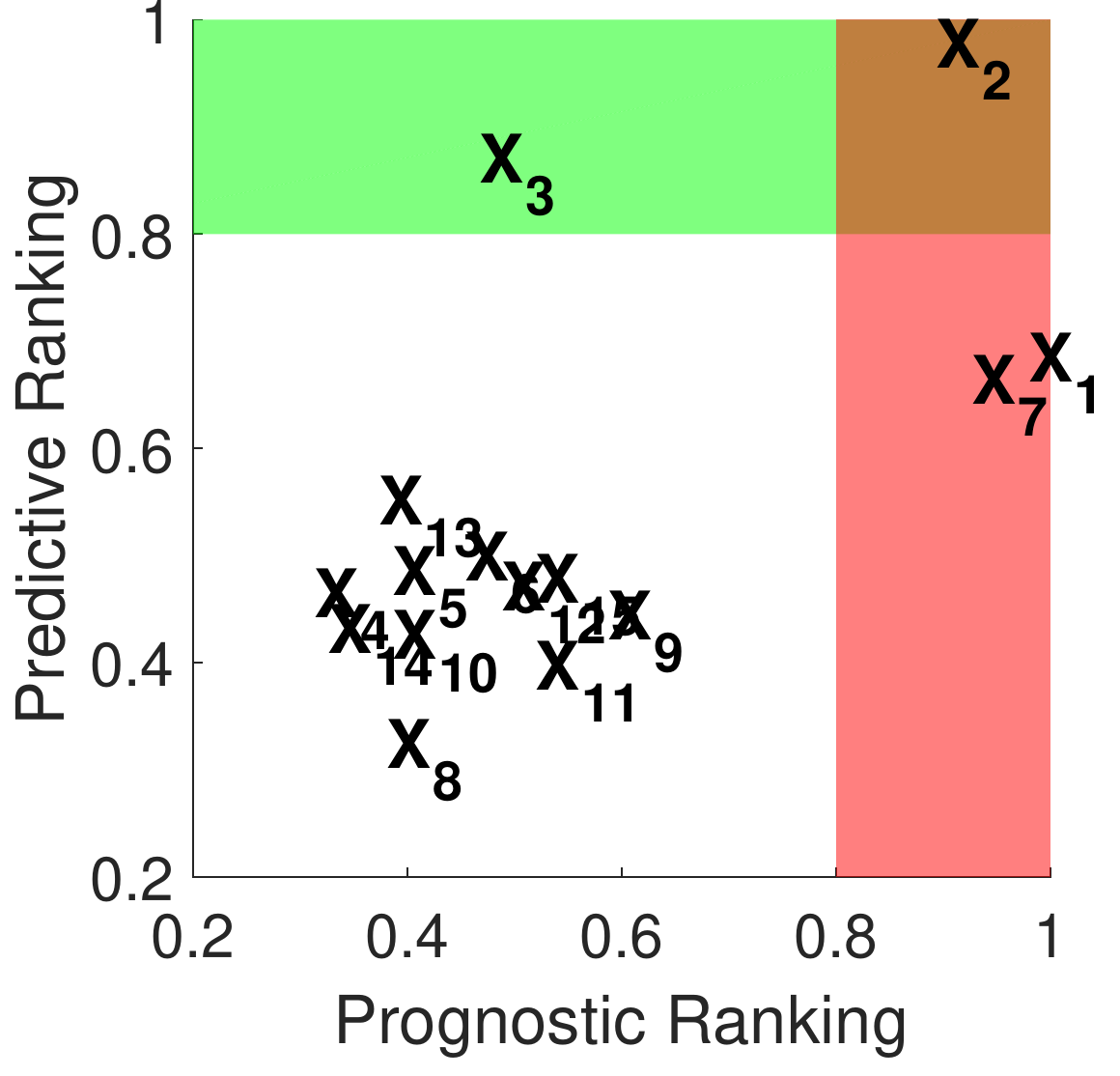}
        \caption{Our $1^{st}$-order approach.}
        \label{fig:MIM}
    \end{subfigure}
    \begin{subfigure}[b]{0.32\textwidth}
        \includegraphics[width=\textwidth]{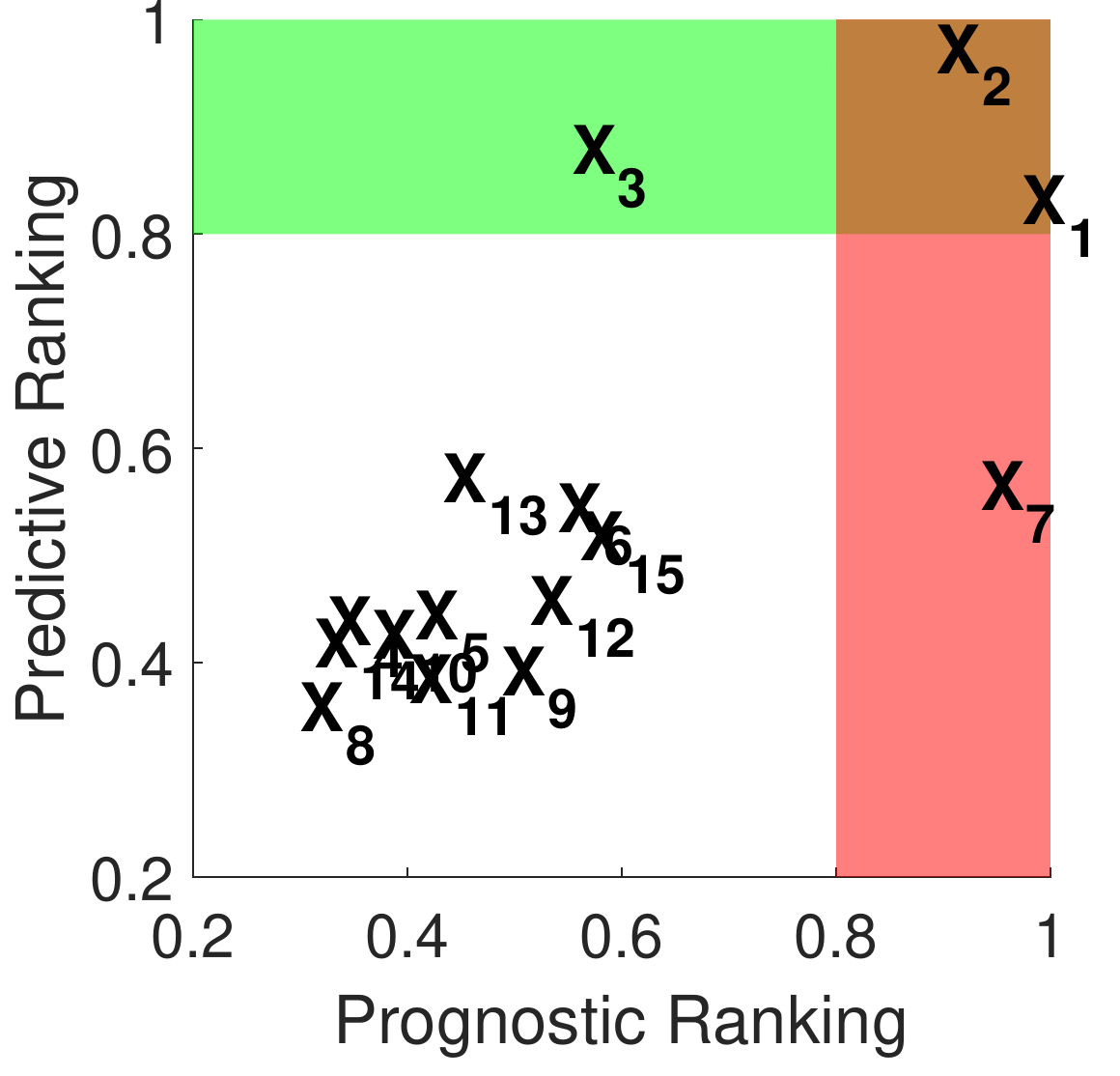}
        \caption{Our $2^{nd}$-order approach.}
        \label{fig:JMI}
    \end{subfigure}
    \caption{{\bf P-P graphs} when: $X_1,X_2$ and $X_3$ are truly predictive, $X_1,X_2$ and $X_7$ are truly prognostic, and the rest nine biomarkers are irrelevant. Note that our our $2^{nd}$-order approximation distinguishes perfectly between predictive and prognostic. }\label{fig:PP}
\end{figure}

Figure \ref{fig:PP} shows three PP-graphs. In the $x$-axis we have the normalised score of each biomarker derived by a prognostic ranking. We normalised scores to take values from $[0,1]$, where $1$ is the score for the most-prognostic biomarker. In the $y$-axis we have the normalised scores for the predictive ranking. The red area (vertical shaded region) represents the top-$k$ prognostic-biomarkers, while the green (horizontal shaded region) the top-$k$ predictive, for these specific PP-graphs we used $k=3,$ which corresponds to the score cut-off value of $(p-k)/p=(15-3)/15=0.80$. The intersection of these two areas -- orange area (top right shaded corner)-- should contain the biomarkers that are both prognostic and predictive.
We plot the average predictive/prognostic rankings over $100$ sample datasets, using Virtual-twins \citep{FosterEtAll2011} and our two approaches suggested in Theorem \ref{thm}. For estimating mutual information, the features were discretized in $4$ equal width bins. As we observe, Virtual-twins \citep{FosterEtAll2011}, tends to push a prognostic biomarker (i.e. $X_7$) into the predictive area --{\em false positive}. The $1^{st}$-order approach classifies $X_1$ only as prognostic and not as predictive --{\em false negative}. While, our $2^{nd}$-order criterion distinguishes perfectly between predictive and prognostic.
\section{Conclusions and Future Work}
In this work we focused on disentangling rankings of the biomarkers that quantify their predictive and their prognostic power. We presented an information-theoretic approach, where we started from a clearly specified objective function and we suggested lower-order approximations. Furthermore, we suggested an efficient estimator for these approximations, by using an empirical Bayes approach to estimate conditional mutual information. Lastly, we introduced a new graphical representation that captures the dynamics of biomarker ranking.
For future work we are planning to apply our methodologies in discovering cardiovascular events in patients undergoing hemodialysis \citep{CardiovascularDS2009}. 
This study contains numerical and categorical covariates. Since discretizing the numerical features it may be a suboptimal solution, we should explore ways of handling them directly. One potential approach is by using the maximal information coefficient \citep{reshef2011detecting}.
Another interesting direction is to improve the interpretability of the PP-graphs. For example, in the $1^{st}$-order approach, instead of plotting the ranking score of each biomarker, we can plot a $p$-value, derived from a univariate testing of whether the biomarker is predictive or prognostic. 

\bibliographystyle{plainnat}
\bibliography{./Bibliography}

\end{document}